\documentclass[letterpaper,journal,twoside]{IEEEtran}

\title{\LARGE \bf
LUMO: Designing Luminous Contact Morphology for Repeatable Whole-Finger Contact Observation
}

\author{
        Dong Ho Kang$^{1}$,
        Youngsu Ko$^{1}$,
        and Luis Sentis$^{1}$

\thanks{$^{1}$ D.H. Kang, Y. Ko, and L. Sentis are with The University of Texas at Austin, Austin, TX, USA (email: {\tt\small dongho@utexas.edu}).
}
}

\usepackage{xcolor}
\usepackage{amsfonts}
\usepackage{amsmath}
\usepackage{amssymb}
\usepackage{graphicx}
\usepackage{tabularx}
\usepackage{booktabs}
\usepackage{dsfont}
\usepackage{multirow}
\usepackage{enumitem}
\usepackage[T1]{fontenc}
\usepackage[utf8]{inputenc}
\usepackage{newtxtext,newtxmath}

\usepackage{cite}

\usepackage{algpseudocode}
\usepackage{algorithm}

\usepackage[hypertexnames=false]{hyperref}

\setlist[itemize]{leftmargin=*}

\newif\ifshowrevisions
\showrevisionsfalse

\ifshowrevisions

  \newcommand{\delrev}[1]{\textcolor{red}{#1}}
\else

  \newcommand{\delrev}[1]{}
\fi

\begin{document}

\maketitle

\begin{abstract}
A low-impedance robot finger reports through joint torque how strongly it is
loaded, but the same torque can arise from a small force near the fingertip or a
large force near the joint. Resolving the force therefore requires knowing where
along the finger contact occurred. LUMO makes that location externally
observable. Embedded LEDs illuminate a compliant silicone pad, and contact
deforms the pad so that light emerging from the finger's side changes in a
pattern set by where the load acts. Because the same structure also carries the
contact load, we optimize its cross-section, including the pad profile, rigid
carrier, and lateral void, for two behaviors at once. Mechanically, the pad
conforms under low preload while the carrier increasingly restricts further
deformation as load rises. Optically, different contact locations produce
separated responses on the finger's side. The search uses rigid--soft contact
simulation, ray tracing, and multi-objective Bayesian optimization. Across two
silicones, six contact locations, and 10- and 30-mm spherical indenters, the
optimized morphologies improve neighboring-location separation relative to
variation from re-establishing contact by \(15\)--\(59\%\). Estimating contact
location from the optical response using the known LED spacing and combining it
with joint torque gives \(1.44~\mathrm{N}\) normal-force MAE over 931 samples.
In a two-finger hand, localized side responses appear on several links
simultaneously during grasps.
\end{abstract}

\section{Introduction}
\label{sec:introduction}

Robotic fingers are commonly designed primarily to position the fingertip,
and contact sensing is typically localized to the distal region~\cite{lepora2026tactile}.  
The palmar surfaces of proximal links, however,
can also participate in manipulation~\cite{reynaerts1994wholefinger,bircher2021complex}. Low-impedance
quasi-direct-drive (QDD) actuation makes contact loads along these links
observable through joint torque~\cite{romero2024eyesight,kang2026plato}. 
This provides load information without distributed contact sensing, but joint torque alone does not uniquely
determine contact location~\cite{pang2021identifying}.

This leaves a gap between where the finger can make useful contact and where
contact can be observed. Distributing sensing over the whole finger requires
additional sensor elements, wiring, and mechanical integration as the
instrumented surface grows~\cite{tomo2018uskin}. External vision avoids
instrumenting each contact region~\cite{kim2023im2contact}, but turns contact
localization into an indirect inference problem whose cues depend on object
geometry, viewpoint, and occlusion.

What if the finger itself were designed to make contact observable away from
the contact interface? Optical tactile sensing has long used light to observe
contact-induced deformation, typically by imaging a compliant contact surface
from behind~\cite{johnson2009retrographic}. Instead of imaging the contact
surface, we design the finger morphology so that contact-induced deformation
changes light transport toward the finger's lateral surfaces. With appropriate
light-source placement and internal geometry, contacts at different longitudinal
locations can therefore produce different optical responses on externally
visible regions, even when the contact interface itself is hidden. Because the
lateral surface extends along the finger, the same observation principle can
cover multiple potential contact locations without requiring a dedicated sensing
element at each site.

\begin{figure}[t]
    \centering
    \includegraphics[width=\linewidth]{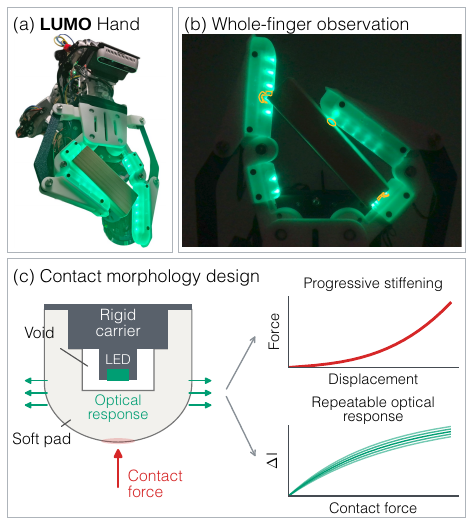}
    \caption{
    \textbf{LUMO concept and integrated hand.}
    (a) LUMO fingers integrated into a force-transparent planar hand.
    (b) Representative whole-finger contact configuration showing externally visible
    illuminated finger regions (green) and contact-facing regions (orange dashed).
    (c) Contact morphology jointly shapes mechanical and optical behavior: the
    compliant pad, rigid carrier, and internal void are designed to promote
    progressive stiffening under increasing load and repeatable spatial optical
    responses across repeated contacts.
}
    \label{fig:fig1}
\end{figure}

This makes the finger's \emph{contact morphology} a central sensing-design
variable. The geometry of the compliant pad, rigid carrier, and internal void
determines how contact forces are transmitted, how the finger deforms, and how
that deformation changes light transport. Contact morphology design therefore
couples mechanical and optical objectives. Mechanically, the interface should
form useful contact while limiting continued deformation as load increases.
Optically, responses from different contact locations should remain spatially
distinguishable and consistent across re-established contacts. Compliance can
improve contact conformation, but soft interfaces can also exhibit
loading--unloading hysteresis and time-dependent relaxation, causing nominally
identical contacts to produce different optical responses
~\cite{wang2026endotac}. A large optical response alone is therefore
insufficient.

To address this coupled design problem, we introduce
\textbf{LUMO (\underline{Lu}minous \underline{M}orphology for Contact
\underline{O}bservation)}. LUMO treats \emph{progressive-constraint contact} as
the mechanical design target: the pad forms a finite contact patch at low
preload while the carrier increasingly limits further deformation as load rises.
Because the optical response arises from contact-induced deformation, we
hypothesize that limiting continued deformation can also reduce optical
variation across independently re-established contacts.

We optimize progressive-constraint contact and spatial optical separation using
contact simulation, ray tracing, and multi-objective Bayesian optimization.
Re-contact repeatability is evaluated physically rather than optimized directly,
since doing so would require multiple independent contact realizations for every
candidate. The morphology-shaped optical response also provides a cue to where
along the finger contact occurs; combined with joint torque, this supplies the
geometry needed for location-aware force estimation. Physical experiments then
evaluate re-contact repeatability, spatial distinguishability, deformation
limiting, and location-aware force estimation.

Our key contributions are:
\begin{itemize}

\item \textbf{Contact-Morphology Co-Design:}
A computational framework that designs the load-bearing finger morphology to
shape both progressive-constraint contact and location-dependent optical
responses on surfaces away from the contact interface.

\item \textbf{Physical Validation:}
Experimental characterization of optimized morphologies across two silicone
materials and two indenter sizes, evaluating spatial distinguishability,
re-contact repeatability, and high-load deformation limiting.

\item \textbf{Location-Aware Force Estimation:}
A causal LED-referenced contact-location estimate that supplies the contact
geometry needed to combine joint torque with normal-force estimation along the
finger.

\end{itemize}

\begin{figure*}[t]
    \centering
    \includegraphics[width=\textwidth]{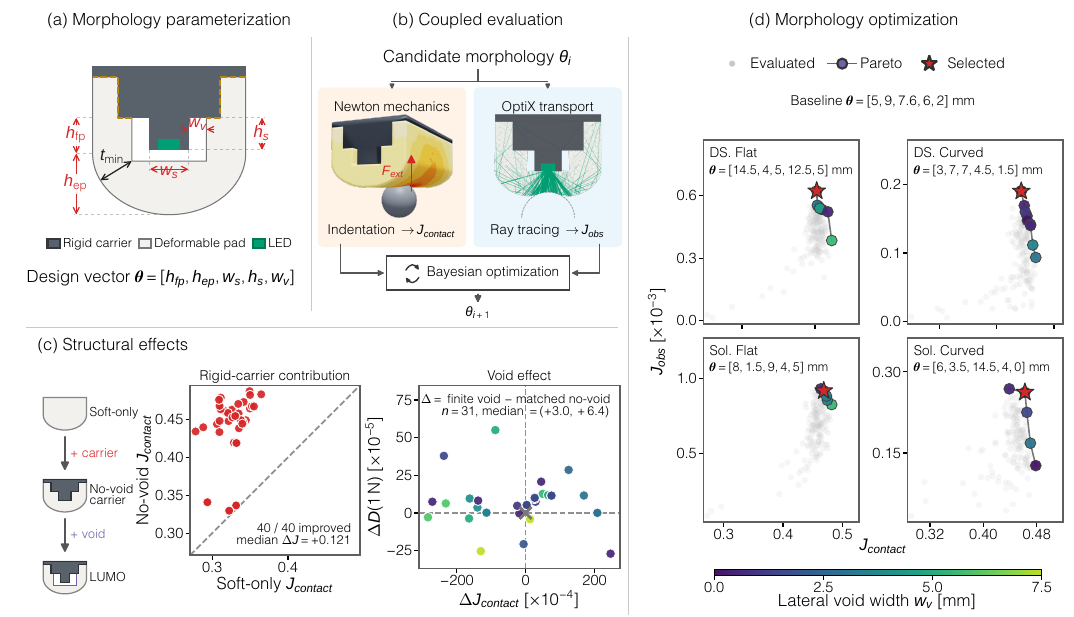}
    \caption{
    \textbf{LUMO morphology-design pipeline.}
    (a) A low-dimensional cross-sectional parameterization defines the compliant
    pad, rigid carrier, internal void, and embedded light-source geometry.
    (b) Each candidate is evaluated by Newton contact simulation and OptiX
    light transport to obtain \(J_{\mathrm{contact}}\) and \(J_{\mathrm{obs}}\),
    which guide Bayesian optimization.
    (c) Matched structural comparisons evaluate the rigid-carrier effect against
    the corresponding soft-only geometry and the lateral-void effect against a
    matched no-void design. For the void comparison, \(\Delta\) denotes the
    finite-void value minus the matched no-void value, and
    \(D(1~\mathrm{N})=\|\widetilde{\mathbf r}_{1\mathrm{N}}\|_2\)
    denotes the unloaded-to-\(1~\mathrm{N}\) optical-state change.
    (d) Evaluated designs and empirical Pareto sets for the four material--loading
    optimization campaigns. Point color indicates lateral void width \(w_v\);
    stars mark the designs selected by Eq.~\eqref{eq:design_selection} by
    maximizing the weaker normalized objective.
    }
    \label{fig:fig2-3}
\end{figure*}

\section{Related Work}
\label{sec:related-work}

\subsection{Whole-Finger Contact Observation}

Without distributed tactile skin, contact location can be inferred from
force/torque or proprioceptive measurements
~\cite{zhou1996contact,manuelli2016contact,wang2020velocity}. These methods
combine interaction measurements with robot geometry using optimization,
filtering, or kinematic constraints. Joint-torque-based inference can become
non-identifiable when distinct contacts produce similar measurements, with
additional ambiguity under multiple contacts
~\cite{manuelli2016contact,pang2021identifying}.
Learning-based approaches can avoid explicit dynamics models or joint-torque
measurements by learning a mapping from proprioceptive observations to surface
contact locations~\cite{liang2021contact}. External RGB-D vision
can instead infer single or multiple contacts without tactile instrumentation
~\cite{kim2023im2contact}. These approaches recover contact from measurements
of the robot or scene. LUMO instead shapes the finger morphology so that
contact itself produces spatially structured cues on an externally observable
surface.

\subsection{Computational Optomechanical Design}

Computational design has been used to shape both the optical and mechanical
response of tactile sensors. Optical simulation has been used to optimize
illumination and optical geometry for photometric reconstruction
~\cite{taylor2022gelslim,agarwal2025pbr}, while elastic-body optimization and
mechanical calibration have improved deformation-based force inference
~\cite{zhang2023elastic,zhao2023insitu}. Coupled mechanical--optical
simulation further enables compliant geometry, stiffness, pad shape, and
illumination to be evaluated before fabrication
~\cite{luu2023simtacls,ma2024simulation}. Multi-objective methods have also
explored trade-offs between sensing and actuation in soft fingers
~\cite{navarro2023design}. LUMO builds on this computational design paradigm
but treats contact progression and spatial optical separation as coupled
objectives of the finger morphology.

\subsection{Contact Morphology Design}

Contact behavior has also been treated as a design objective for end-effector
geometry and compliance. Finger profiles have been synthesized from prescribed
contact constraints~\cite{rodriguez2013effector}, and fingertip surfaces have
been optimized for representative local contact geometries
~\cite{song2018fingertip}. Mechanical models relate fingertip deformation to
contact force~\cite{inoue2006elastic}, while rigid reinforcement can suppress
global deformation while preserving local compliance
~\cite{kang2026plato}. Internal interactions such as layer jamming and
self-contact have likewise been designed to regulate deformability and force
generation~\cite{elgeneidy2020stiffening,navez2024selfcontacts}. These works
shape how the interface mechanically responds to contact. LUMO extends this
view by designing the same morphology for progressive-constraint contact and
spatially distinguishable optical responses.

\section{Methodology}
\label{sec:methodology}

\begin{table*}[t]
    \centering
    \caption{\textbf{Simulation and optimization settings.}}
    \label{tab:method_settings}
    \vspace{-1.5mm}
    \footnotesize
    \renewcommand{\arraystretch}{0.90}

    \begin{minipage}[t]{0.49\textwidth}
    \centering
    \begin{tabularx}{\linewidth}{l l X}
        \toprule
        Group & Setting & Value \\
        \midrule

        \multirow{3}{*}{Morphology}
        & Design vector
        & \(\boldsymbol{\theta}
        =[h_{\mathrm{fp}},h_{\mathrm{ep}},w_s,h_s,w_v]^{\mathsf T}\) \\

        & Bounds [mm]
        & \([2,1,4,2,0]^{\mathsf T}
        \leq \boldsymbol{\theta} \leq
        [29,20,15,15,7.5]^{\mathsf T}\) \\

        & Constraints
        & \(h_{\mathrm{fp}}+h_{\mathrm{ep}}\leq20\),
          \(w_s+2w_v\leq19.5\),
          \(t_{\min}\geq5~\mathrm{mm}\);
          \(0.5~\mathrm{mm}\) lattice \\

        \midrule

        \multirow{3}{*}{Mechanics}
        & Material
        & DS: \(\rho=1070~\mathrm{kg/m^3}\),
          \(\mu=106~\mathrm{kPa}\),
          \(\lambda=10.494~\mathrm{MPa}\);
          Sol.: \(\rho=990~\mathrm{kg/m^3}\),
          \(\mu=98.5~\mathrm{kPa}\),
          \(\lambda=9.750~\mathrm{MPa}\) \\

        & Numerical
        & \(1~\mathrm{mm}\) mesh,
          \(0.01~\mathrm{s}\) step,
          10 VBD iterations,
          \(\eta=10~\mathrm{Pa\,s}\) \\

        & Loading
        & \(5~\mathrm{mm/s}\);
          \(1,2,5,10,15~\mathrm{N}\);
          \(1~\mathrm{s}\) hold \\

        \bottomrule
    \end{tabularx}
    \end{minipage}
    \hfill
    \begin{minipage}[t]{0.49\textwidth}
    \centering
    \begin{tabularx}{\linewidth}{l l X}
        \toprule
        Group & Setting & Value \\
        \midrule

        \multirow{2}{*}{Optics}
        & Geometry
        & 5 LEDs, \(p_{\mathrm{LED}}=11~\mathrm{mm}\);
          \(B=11\), \(Y\in[-27.5,27.5]~\mathrm{mm}\),
          \(5~\mathrm{mm}\) bins \\

        & Material
        & DS: \(n=1.4348\),
          \(\mu_{\mathrm{ext}}=71.3801~\mathrm{m^{-1}}\);
          Sol.: \(n=1.4100\),
          \(\mu_{\mathrm{ext}}=14.5063~\mathrm{m^{-1}}\) \\

        \midrule

        \multirow{2}{*}{Scenarios}
        & Geometry
        & \(d\in\{10,15,20,30\}~\mathrm{mm}\);
          \(y/p_{\mathrm{LED}}
          \in\{-2,-1,-1/2,0,1/2,1,2\}\) \\

        & Angle
        & Flat: \(0^\circ\);
          Curved:
          \(\{-30,-15,0,15,30\}^\circ\) \\

        \midrule

        \multirow{2}{*}{Optimization}
        & Acquisition
        & qLogNEHVI; 5 heuristically selected reference designs
          + 8 Sobol initial evaluations \\

        & Search
        & 256 feasible candidates per iteration;
          160 successful evaluations per campaign \\

        \bottomrule
    \end{tabularx}
    \end{minipage}

    \vspace{-2mm}
\end{table*}

LUMO searches a low-dimensional contact-morphology space using coupled
mechanical and optical simulation. The resulting objectives,
\(J_{\mathrm{contact}}\) and \(J_{\mathrm{obs}}\), guide multi-objective
Bayesian optimization (Fig.~\ref{fig:fig2-3}).

\subsection{Morphology Parameterization}
\label{sec:morphology_parameterization}

Each morphology is defined by
\(\boldsymbol{\theta}
=[h_{\mathrm{fp}},h_{\mathrm{ep}},w_s,h_s,w_v]^{\mathsf T}\),
where \(h_{\mathrm{fp}}\) and \(h_{\mathrm{ep}}\) define the flat and
elliptical pad regions, \(w_s\) and \(h_s\) the internal stem, and \(w_v\)
the lateral void width [Fig.~\ref{fig:fig2-3}(a)]. The resulting
cross-section defines the compliant pad, rigid carrier, and embedded
light-source geometry, with \(t_{\min}\) denoting the minimum silicone
thickness. Designs violating the constraints in
Table~\ref{tab:method_settings} or producing invalid cross-sections are
discarded.

\subsection{Coupled Mechanical and Optical Evaluation}
\label{sec:coupled_evaluation}

Each candidate \(\boldsymbol{\theta}_i\) is evaluated under contact scenario
\(s=(d,\alpha,y)\), comprising spherical-indenter diameter \(d\), approach
angle \(\alpha\) relative to the pad normal, and longitudinal contact location
\(y\). At load state \(k\), mechanical simulation returns the deformed
configuration \(\mathbf{X}_{s,k}\), indenter-contact set
\(\mathcal{C}_{s,k}\), reaction force \(F_{s,k}\), and indentation depth
\(\delta_{s,k}\); ray tracing maps \(\mathbf{X}_{s,k}\) to spatial optical
response \(\mathbf{r}_{s,k}\).

\subsubsection{Contact Mechanics}

Contact mechanics are simulated in NVIDIA Newton using its GPU-based Vertex
Block Descent (VBD) solver~\cite{newton2025,chen2024vbd}. The compliant pad
is tetrahedralized and modeled using Newton's damped Neo-Hookean material
model with the Lam\'e parameters and damping coefficient listed in
Table~\ref{tab:method_settings}. The rigid carrier is fixed,
bonded-interface vertices are kinematically constrained, and the remaining
pad--carrier interaction uses penalty contact with signed-distance-field
full-surface rigid--soft collision detection. Silicone self-contact is
disabled.

For each scenario, a rigid spherical indenter advances along the prescribed
approach direction. The reaction force is obtained from Newton's
contact-wrench output and projected onto this direction to obtain
\(F_{s,k}\). At each prescribed load, the indenter is held for
\(1~\mathrm{s}\) before the mechanical state is recorded for contact and
optical evaluation; \(k\) indexes these loads in increasing order.

\subsubsection{Optical Response}

For each mechanical state \(\mathbf{X}_{s,k}\), the silicone geometry is
updated from the Newton vertices while the rigid carrier and stem-mounted LEDs
remain fixed. NVIDIA OptiX~\cite{parker2010optix} evaluates light transport
with finite diffuse emitters, Fresnel reflection and refraction, total
internal reflection, Beer--Lambert attenuation, and diffuse carrier
reflection~\cite{hecht2017optics}.

The optical parameters in Table~\ref{tab:method_settings} use the
manufacturer-specified Solaris refractive index~\cite{smoothon_solaris}, a
reported refractive index for transparent PDMS for Dragon
Skin~\cite{schneider2009pdms}, and extinction coefficients converted from
reported visible-light PDMS propagation losses
~\cite{wang2019flexible,azmayeshfard2010pdms}. The effective extinction
coefficient therefore combines scattering and absorption.

The externally visible finger surface is divided into \(B\) fixed
longitudinal regions, with optical response
\begin{equation}
\mathbf{r}_{s,k}
=
\begin{bmatrix}
P^{(1)}_{s,k} &
P^{(2)}_{s,k} &
\cdots &
P^{(B)}_{s,k}
\end{bmatrix}^{\mathsf T},
\label{eq:optical_response}
\end{equation}
where \(P^{(b)}_{s,k}\) is the power emerging from region \(b\).
Contributions from all LEDs are combined using the same deterministic
ray-sampling pattern across morphologies and mechanical states.

\subsection{Mechanical and Optical Objectives}
\label{sec:objectives}

Each morphology is evaluated by progressive-constraint contact,
\(J_{\mathrm{contact}}\), and spatial optical separation,
\(J_{\mathrm{obs}}\):
\begin{equation}
\underset{\boldsymbol{\theta}\in\Theta}{\mathrm{maximize}}
\quad
\left(
J_{\mathrm{contact}}(\boldsymbol{\theta}),
J_{\mathrm{obs}}(\boldsymbol{\theta})
\right).
\label{eq:multiobjective}
\end{equation}

\subsubsection{Progressive-Constraint Contact}

The mechanical objective evaluates finite low-load patch formation,
retention of the contacted surface as load increases, and progressive
stiffening. For contact scenario \(s\),
\begin{equation}
\begin{aligned}
q_{\mathrm{form},s}
&=
\min\!\left(
1,
\sqrt{
\frac{
A(\mathcal{C}_{s,2};\mathbf{X}_{s,2})
}{
\pi R_s^2
}}
\right),\\
q_{\mathrm{stable},s}
&=
\frac{
A_0(\mathcal{C}_{s,2}\cap\mathcal{C}_{s,4})
}{
A_0(\mathcal{C}_{s,2}\cup\mathcal{C}_{s,4})
},\\
q_{\mathrm{stiff},s}
&=
\operatorname{clip}_{[0,1]}
\left(
1-\frac{\kappa_{\mathrm{early},s}}
{\kappa_{\mathrm{late},s}}
\right),
\end{aligned}
\label{eq:contact_terms}
\end{equation}
where \(R_s\) is the indenter radius,
\(A(\cdot;\mathbf{X})\) measures deformed contact area, and
\(A_0(\cdot)\) measures the corresponding reference-surface area.
\(q_{\mathrm{form}}\) evaluates patch formation at \(2~\mathrm{N}\), while
\(q_{\mathrm{stable}}\) measures retention of the contacted surface from
\(2\) to \(10~\mathrm{N}\). The early and late secant stiffnesses are
\begin{equation}
\kappa_{\mathrm{early},s}
=
\frac{F_{s,2}-F_{s,1}}
{\delta_{s,2}-\delta_{s,1}},
\qquad
\kappa_{\mathrm{late},s}
=
\frac{F_{s,4}-F_{s,3}}
{\delta_{s,4}-\delta_{s,3}}.
\label{eq:secant_stiffness}
\end{equation}
These secants span \(1\)--\(2~\mathrm{N}\) and \(5\)--\(10~\mathrm{N}\),
respectively, so \(q_{\mathrm{stiff}}\) rewards increasing stiffness with
load.

The three terms are combined by their geometric mean, with the minimum over
the prescribed scenarios:
\begin{equation}
J_{\mathrm{contact}}
=
\min_s
\left(
q_{\mathrm{form},s}
q_{\mathrm{stable},s}
q_{\mathrm{stiff},s}
\right)^{1/3}.
\label{eq:j_contact}
\end{equation}
The geometric mean penalizes weak individual properties, while the outer
minimum selects the worst tested scenario. \(J_{\mathrm{contact}}\) scores
the contact progression hypothesized to improve re-contact repeatability;
repeatability itself is evaluated experimentally. The \(15~\mathrm{N}\)
state contributes to the optical objective but not to
\(J_{\mathrm{contact}}\).

\subsubsection{Spatial Optical Separation}

Each optical response is referenced to the unloaded state and normalized by
the total modeled emitted power:
\begin{equation}
\widetilde{\mathbf{r}}_{s,k}
=
\frac{
\mathbf{r}_{s,k}-\mathbf{r}_0
}{
P_{\mathrm{emit}}
}.
\label{eq:normalized_optical_response}
\end{equation}

Responses are compared only at matched indenter diameter, approach angle,
and load:
\begin{equation}
J_{\mathrm{obs}}
=
\min_{d,\alpha,k}
\;
\min_{y_a\neq y_b}
\left\|
\widetilde{\mathbf{r}}_{(d,\alpha,y_a),k}
-
\widetilde{\mathbf{r}}_{(d,\alpha,y_b),k}
\right\|_2 .
\label{eq:j_obs}
\end{equation}
Thus, \(J_{\mathrm{obs}}\) is the worst-case optical-state separation among
the tested contact locations under matched interaction conditions.

\begin{figure}[t]
    \centering
    \includegraphics[width=0.45\textwidth]{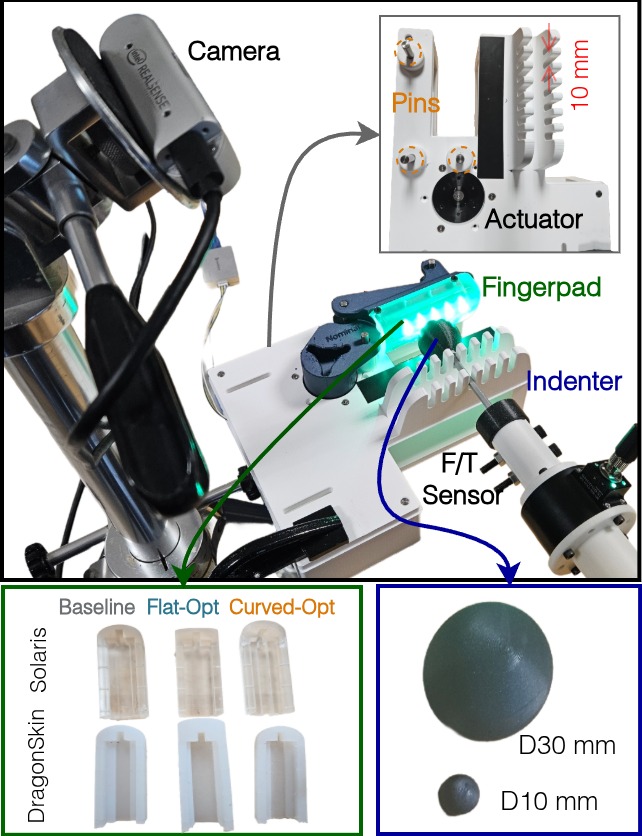}
    \caption{
        \textbf{Experimental setup.}
        A Bota Rokubi force/torque sensor applies contact through 10- and
        30-mm spherical indenters at six indexed longitudinal locations spaced
        by \(10~\mathrm{mm}\). An Intel RealSense D435 observes the finger.
        Six specimens span three morphologies and two silicone materials.
        Auto exposure was disabled, with exposure and gain fixed at device
        settings 10 and 64, respectively.
    }
    \label{fig:fig4}
\end{figure}

\begin{figure*}[t]
    \centering
    \includegraphics[width=\textwidth]{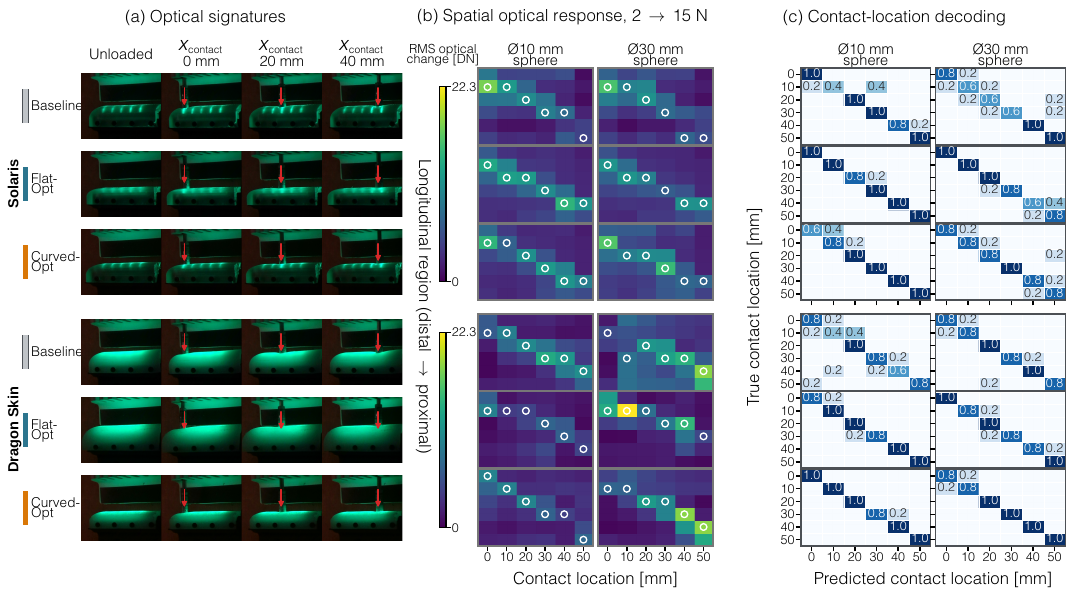}
    \caption{
        \textbf{Spatial optical response and discrete contact-location decoding.}
        Baseline, Flat-Opt, and Curved-Opt morphologies are evaluated in Solaris
        and Dragon Skin at six longitudinal locations spaced by \(10~\mathrm{mm}\)
        using 10- and 30-mm spherical indenters.
        (a) Representative unloaded and loaded images at selected contact locations.
        (b) Regional RMS green-channel optical change, in camera digital numbers
        (DN), from 2 to \(15~\mathrm{N}\), summarized over five independent
        re-contacts; open circles indicate the region of maximum response.
        (c) Leave-one-repetition-out six-class decoding using the signed
        2-to-\(5~\mathrm{N}\) regional response and nearest-location templates.
        Each confusion matrix contains 30 held-out predictions, with predictions
        restricted to the six tested locations.
    }
    \label{fig:fig5}
\end{figure*}

\subsection{Bayesian Optimization and Design Selection}
\label{sec:optimization}

We solve Eq.~\eqref{eq:multiobjective} sequentially using logarithmic noisy
expected hypervolume improvement (qLogNEHVI) implemented in Ax. Each campaign
is initialized with five heuristically selected reference designs and eight
Sobol samples, and only successful evaluations count toward the campaign
budget (Table~\ref{tab:method_settings}). The \emph{Flat} ensemble uses
pad-normal contact, whereas the \emph{Curved} ensemble spans the approach
angles in Table~\ref{tab:method_settings}. Optimizing each ensemble separately
for Dragon Skin (DS.) and Solaris (Sol.) yields four campaigns, whose
nondominated evaluated designs define the empirical Pareto sets in
Fig.~\ref{fig:fig2-3}(d).

Within each campaign, each objective is normalized by its maximum over the
evaluated designs, and one design is selected by maximizing the weaker
normalized objective:
\begin{equation}
\begin{aligned}
\widetilde{J}_{\mathrm{contact},i}
&=
\frac{J_{\mathrm{contact},i}}
{\max_j J_{\mathrm{contact},j}},
\qquad
\widetilde{J}_{\mathrm{obs},i}
=
\frac{J_{\mathrm{obs},i}}
{\max_j J_{\mathrm{obs},j}},
\\
i^\star
&=
\arg\max_i
\min\!\left(
\widetilde{J}_{\mathrm{contact},i},
\widetilde{J}_{\mathrm{obs},i}
\right).
\end{aligned}
\label{eq:design_selection}
\end{equation}
The selected designs are denoted Flat-Opt and Curved-Opt according to their
loading ensemble.

\section{Evaluation}
\label{sec:experiments}

\subsection{Morphology Design Study}
\label{sec:design-study}

Matched comparisons isolate the structural roles of the rigid carrier and
lateral void [Fig.~\ref{fig:fig2-3}(c)]. Adding the carrier increases
\(J_{\mathrm{contact}}\) in all 40 matched designs
(median \(\Delta J_{\mathrm{contact}}=0.121\)), confirming its contribution
to the simulated contact objective. A finite lateral void produces
morphology-dependent changes in both \(J_{\mathrm{contact}}\) and the
unloaded-to-\(1~\mathrm{N}\) optical response change
\(\Delta D(1~\mathrm{N})\), motivating \(w_v\) as an optimization variable.

The four optimization campaigns yield the empirical Pareto sets in
Fig.~\ref{fig:fig2-3}(d). Flat-Opt and Curved-Opt are selected using
Eq.~\eqref{eq:design_selection} and fabricated, together with the nominal
Baseline, for physical evaluation.

\begin{figure*}[t]
    \centering
    \includegraphics[width=\textwidth]{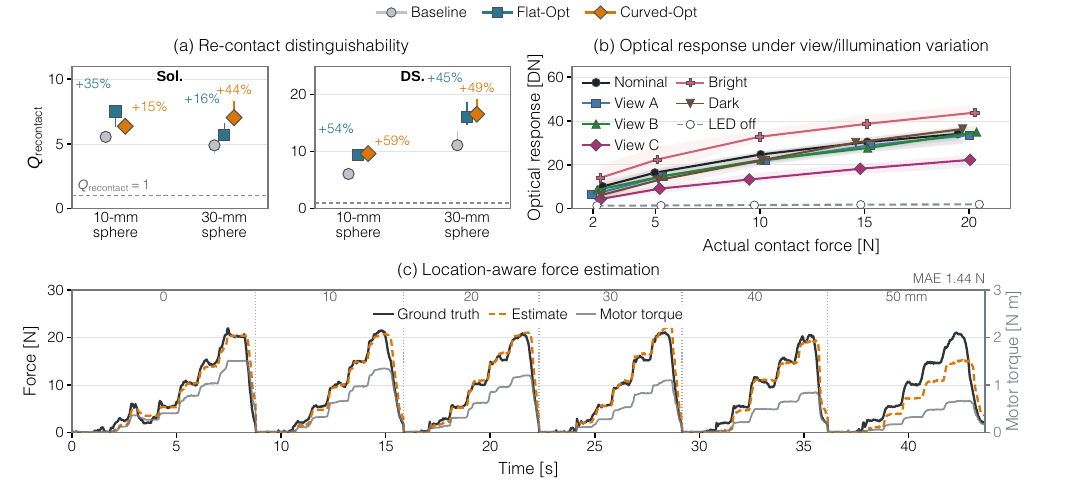}
    \caption{
        \textbf{Re-contact distinguishability, imaging variation, and
        location-aware force estimation.}
        (a) Re-contact distinguishability
        \(Q_{\mathrm{recontact}}
        =D_{\mathrm{neighbor}}/W_{\mathrm{recontact}}\)
        for Solaris and Dragon Skin using 10- and 30-mm spherical indenters.
        Whiskers show the minimum and maximum values obtained after deleting one
        re-contact and recomputing the metric; \(Q_{\mathrm{recontact}}=1\)
        is shown as a scale reference.
        (b) Force-dependent optical response of Solaris Curved-Opt using the
        10-mm indenter at three longitudinal locations while camera height and
        pitch were manually varied, together with changes in ambient illumination.
        Shaded regions span the tested locations, and LED-off provides a matched
        control.
        (c) Location-aware proprioceptive force estimation across six
        longitudinal contact locations. The causal LED-referenced estimate
        achieves \(1.44~\mathrm{N}\) MAE over 931 contact samples from six
        loading sequences; external force/torque measurements provide ground
        truth and motor torque is shown on the secondary axis.
    }
    \label{fig:fig6}
\end{figure*}

\begin{figure*}[t]
    \centering
    \includegraphics[width=\textwidth]{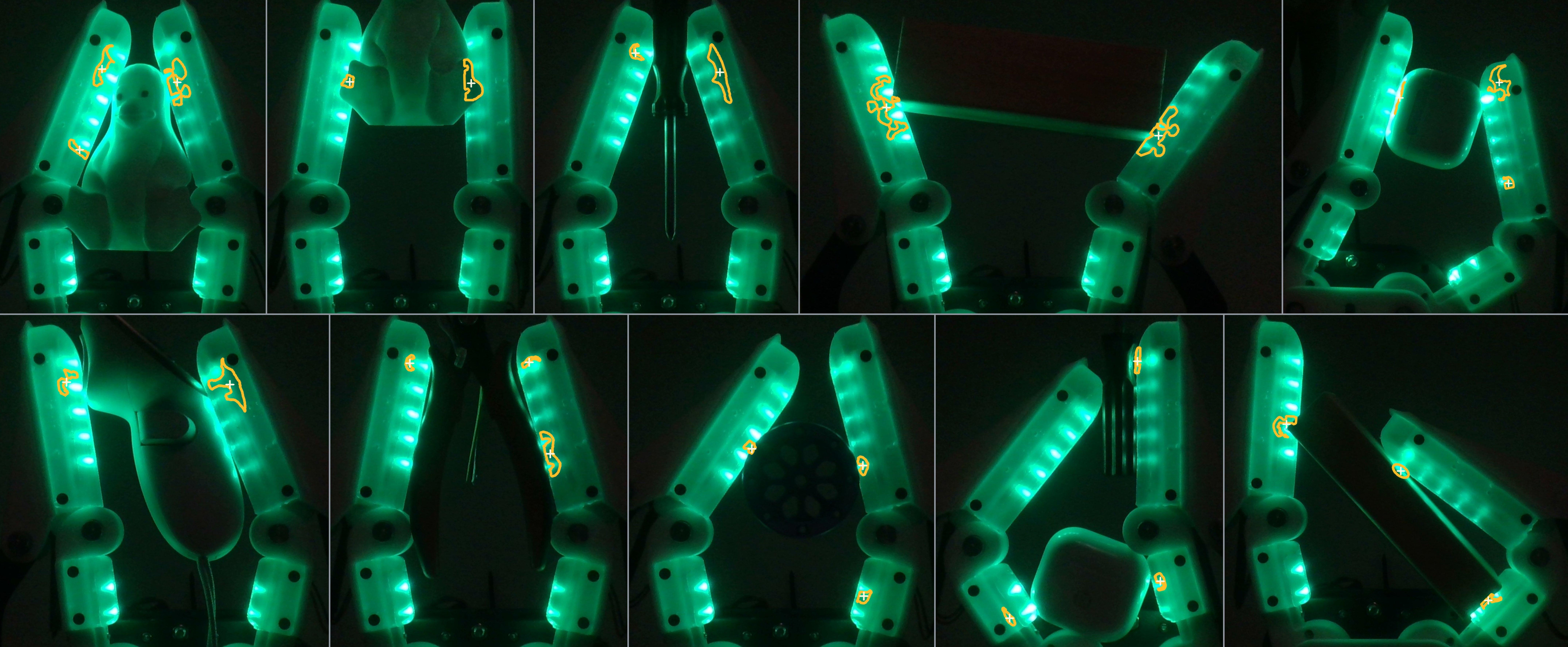}
    \caption{
        \textbf{Qualitative multi-contact optical responses in the integrated
        4-DoF hand.}
        Yellow contours indicate detected optical-change regions relative to
        the nearest unloaded reference in joint-configuration space, and
        crosses mark their residual-weighted centroids. Multiple localized
        responses are identified simultaneously across distal and proximal
        finger links under different grasp configurations.
    }
    \label{fig:fig7}
\end{figure*}

\subsection{Spatial Distinguishability and Re-Contact Variability}
\label{sec:recontact}

Baseline, Flat-Opt, and Curved-Opt specimens are evaluated at six longitudinal
locations using 10- and 30-mm spherical indenters (Fig.~\ref{fig:fig4}).
At each location, five independent re-contacts are loaded to
\(2\), \(5\), \(10\), and \(15~\mathrm{N}\), with full unloading between
contacts and a \(1~\mathrm{s}\) hold at each load. Images are reduced to
longitudinal green-channel profiles after subtracting the unloaded response.

For location \(i\) and re-contact \(r\), \(\mathbf{p}_{i,r}\) denotes the
longitudinal profile of intensity--force slopes over the loading sequence.
We define
\begin{equation}
\begin{aligned}
D_{\mathrm{neighbor}}
&=\operatorname*{med}_{i}
\operatorname{RMS}\!\left(
\bar{\mathbf p}_{i+1}-\bar{\mathbf p}_{i}\right),\\
W_{\mathrm{recontact}}
&=\operatorname*{med}_{i,r}
\operatorname{RMS}\!\left(
\mathbf p_{i,r}-\bar{\mathbf p}_{i}\right),\\
Q_{\mathrm{recontact}}
&=D_{\mathrm{neighbor}}/W_{\mathrm{recontact}},
\end{aligned}
\label{eq:recontact_metrics}
\end{equation}
where \(\bar{\mathbf p}_{i}\) is the pointwise median across re-contacts.
\(D_{\mathrm{neighbor}}\) measures separation between neighboring
location-dependent responses, whereas \(W_{\mathrm{recontact}}\) measures
variation introduced by independently re-establishing contact.
Their ratio therefore measures spatial distinguishability relative to
re-contact variability.

As a complementary discrete test, the signed 2-to-\(5~\mathrm{N}\) regional
response is decoded using nearest-location templates under
leave-one-repetition-out evaluation. Predictions are restricted to the six
tested locations, so this test measures discrete separability rather than
continuous localization.

Both optimized morphologies increase \(Q_{\mathrm{recontact}}\) over Baseline
for every tested material--indenter combination, by 15--44\% for Solaris and
45--59\% for Dragon Skin [Fig.~\ref{fig:fig6}(a)]. Thus, optimization
improves location-dependent separation relative to re-contact variation,
rather than merely increasing optical response magnitude.

\begin{table*}[t]
    \centering
    \caption{\textbf{Baseline-relative response changes under repeated contact.}}
    \label{tab:physical_summary}
    \vspace{-1.5mm}
    \small
    \setlength{\tabcolsep}{5.0pt}
    \renewcommand{\arraystretch}{0.88}

    \begin{tabular}{llccc@{\hspace{10pt}}ccc}
        \toprule
        & &
        \multicolumn{3}{c}{Optical Response [\%]} &
        \multicolumn{3}{c}{Progressive-Constraint Gain [\%]} \\[-1pt]

        Material &
        Morphology &
        \(W_{\mathrm{cycle}}\) reduction &
        \(W_{\mathrm{recontact}}\) reduction &
        \(D_{\mathrm{neighbor}}\) gain &
        \(-30^\circ\) &
        \(0^\circ\) &
        \(+30^\circ\) \\
        \midrule

        \multirow{2}{*}{Sol.}
        & Flat-Opt
        & +30.2 & +32.5 & -8.6
        & +0.3 & +19.8 & +3.0 \\

        & Curved-Opt
        & -50.3 & +16.7 & -4.6
        & +19.3 & +21.0 & +20.1 \\

        \midrule

        \multirow{2}{*}{DS.}
        & Flat-Opt
        & +20.3 & +41.9 & -10.4
        & -40.6 & +11.5 & -38.1 \\

        & Curved-Opt
        & -9.6 & +33.2 & +5.9
        & +13.9 & +19.9 & +14.3 \\

        \bottomrule
    \end{tabular}
    \vspace{-2mm}
\end{table*}

\subsection{Contact Cycling and Approach-Angle Response}
\label{sec:cycling}

To distinguish variation within maintained contact from variation after
re-contact, the 10-mm indenter is cycled five times between \(2\) and
\(15~\mathrm{N}\) without breaking contact at locations 1, 3, and 5 and
\(\alpha\in\{-30^\circ,0^\circ,+30^\circ\}\). Three independently
re-established runs are collected per location--angle condition.

For these runs, \(W_{\mathrm{cycle}}\) measures matched-force variation across
cycles within maintained contact, while \(W_{\mathrm{recontact}}\) and
\(D_{\mathrm{neighbor}}\) measure variation across re-established runs and
neighboring-location separation, respectively. Table~\ref{tab:physical_summary}
reports baseline-relative reductions \(100(1-W/W_{\mathrm{base}})\) and
separation gains \(100(D/D_{\mathrm{base}}-1)\).

Re-contact variability decreases for every optimized morphology
(\(16.7\)--\(41.9\%\)), whereas changes in \(D_{\mathrm{neighbor}}\) are
mixed (\(-10.4\) to \(+5.9\%\)). Because \(Q_{\mathrm{recontact}}\) is
defined relative to re-contact variability, reducing
\(W_{\mathrm{recontact}}\) can improve spatial distinguishability even when
\(D_{\mathrm{neighbor}}\) does not increase. The mixed
\(W_{\mathrm{cycle}}\) results further indicate that repeatability within
maintained contact and after re-contact are distinct properties.

High-load deformation is evaluated from the projected displacement of a
marker attached to the rigid indenter, measured from force/torque-detected
contact onset to \(15~\mathrm{N}\). Progressive-Constraint Gain is the
baseline-relative reduction in this displacement. Curved-Opt yields positive
gain at every tested angle in both materials, whereas Flat-Opt is
angle-dependent, supporting the targeted high-load deformation-limiting
behavior.

\subsection{Imaging Variation}
\label{sec:imaging-variation}

Imaging sensitivity is evaluated with Solaris Curved-Opt using the 10-mm
indenter at locations 1, 3, and 5. Loads of \(2\), \(5\), \(10\), \(15\),
and \(20~\mathrm{N}\) are applied under varied camera viewpoints and ambient
illumination, together with a matched LED-off control. The force-dependent
optical response remains ordered under the tested conditions despite changes
in absolute magnitude, whereas the LED-off response is substantially weaker
[Fig.~\ref{fig:fig6}(b)]. This tests persistence of the optical cue under the
tested imaging variations, not viewpoint- or illumination-invariant
localization.

\subsection{Location-Aware Proprioceptive Force Estimation}
\label{sec:force-estimation}

The nearest-template decoder in Sec.~\ref{sec:recontact} is used only to
evaluate discrete separability. For force estimation, we instead use a causal
contact-location estimate from the unloaded-relative optical response.
Solaris Curved-Opt is operated under local PD position control while the
10-mm indenter applies \(2\), \(5\), \(10\), \(15\), and \(20~\mathrm{N}\)
at all six contact locations.

The positive green-channel response is reduced to a longitudinal profile,
filtered using only current and past frames, and its strongest
noise-thresholded peak is mapped to the finger coordinate using the five
detected LED positions and their known \(11~\mathrm{mm}\) spacing.
No labeled contact locations are used.

For the single-joint planar configuration, the contact Jacobian reduces to a
moment arm, so the estimated location provides the geometry needed to convert
joint torque into normal contact force. Motor torque is obtained from actuator
current using the calibrated torque constant. Across 931 contact samples from
six loading sequences, one per contact location, the causal estimate achieves
\(1.44~\mathrm{N}\) MAE. Processing requires \(0.86~\mathrm{ms}\) per frame
at the median (\(0.94~\mathrm{ms}\), 95th percentile) on an Intel Core
i7-9700K CPU, excluding image decoding.

Using the same logs, a fixed moment arm at the center of the tested range
gives \(2.72~\mathrm{N}\) MAE, whereas the known jig locations give
\(1.29~\mathrm{N}\). The optical location estimate therefore recovers most
of the error gap between a fixed moment arm and known contact geometry.
The remaining error includes torque estimation, moment-arm modeling, force
direction, and temporal alignment.

\subsection{Multi-Contact Grasp Responses in the Integrated Hand}
\label{sec:integrated-hand}

We finally examine whether localized optical responses remain observable during
multi-contact grasps with the integrated 4-DoF hand. An unloaded reference
library was collected by sampling several finger configurations during free
motion across the workspace. For each grasp, the nearest reference in the
corresponding finger's joint-configuration space is selected, and the distal
and proximal links are registered independently using their mounting-hole
geometry.

After compensating for global and slowly varying green-channel intensity
differences, the absolute residual between the grasp and unloaded images is
used to identify localized optical-change regions along each link. Prominent
longitudinal responses are segmented into connected regions, and each region
is represented by its residual-weighted centroid. The procedure requires
neither labeled contact locations nor an object model and can identify multiple
response regions simultaneously across the four finger links.

\section{Conclusion}

LUMO treats whole-finger contact observation as a contact-morphology design
problem. The physical results show that optimized morphologies improve
consistency across independently re-established contacts while retaining
spatially distinguishable optical responses. These contact-location cues also
provide the geometric information needed to interpret joint torque as contact
load along the finger. Integrated-hand demonstrations further show that
localized optical responses remain observable across multiple finger links
during multi-contact grasps.

The experiments indicate that a strong optical response alone is insufficient
for reliable spatial observation. Contact morphology must also preserve
location-dependent structure and consistency across repeated contact. This
coupling between mechanical contact behavior and optical response is the main
design principle underlying LUMO.

The quantitative evaluation is limited to single contacts, controlled
longitudinal locations, and the tested loading and imaging conditions. The
coupled simulation is used to rank candidate morphologies rather than to
predict absolute measured responses; quantitative simulation-to-experiment
correspondence would require evaluating a broader set of designs across the
Pareto front. The camera-based implementation also depends on viewing geometry,
as off-normal views and occlusion can reduce visibility of the illuminated
finger surfaces. The camera should therefore be viewed as one readout of the
contact-dependent light transport produced by the finger morphology rather
than as an essential element of LUMO. Future work will examine alternative
optical readouts, such as photodiodes, that may reduce dependence on external
viewing geometry and camera frame rate.



\renewcommand{\baselinestretch}{1.00}
\footnotesize
\bibliographystyle{IEEEtran}
\bibliography{references}


\end{document}